\documentclass{article}

\usepackage{PRIMEarxiv}

\usepackage[utf8]{inputenc} % allow utf-8 input
\usepackage[T1]{fontenc}    % use 8-bit T1 fonts
\usepackage{hyperref}       % hyperlinks
\usepackage{url}            % simple URL typesetting
\usepackage{booktabs}       % professional-quality tables
\usepackage{amsfonts}       % blackboard math symbols
\usepackage{nicefrac}       % compact symbols for 1/2, etc.
\usepackage{microtype}      % microtypography
\usepackage{lipsum}
\usepackage{fancyhdr}       % header
\usepackage{graphicx}       % graphics
\usepackage[normalem]{ulem} % strike through text 
\usepackage{amsmath}
\usepackage{amssymb}
\usepackage{multirow}
\usepackage{array}
\usepackage{booktabs}
\usepackage{geometry}
\usepackage{listings}
\usepackage{xcolor}
\usepackage{todonotes}
\usepackage{booktabs}
\usepackage{subcaption}
\usepackage[style=alphabetic]{biblatex}
\usepackage{appendix}

\lstdefinestyle{prompt}{
    backgroundcolor=\color{gray!10},
    basicstyle=\ttfamily,
    frame=single,
    breaklines=true,
    captionpos=b,
    breakindent=0pt,
}
\lstdefinestyle{lineartext}{
    backgroundcolor=\color{gray!10},
    basicstyle=\ttfamily\tiny,
    frame=single,
    breaklines=true,
    breakindent=0pt,
}

\title{
    Large Language Models for Page Stream Segmentation
}

\usepackage{authblk}

\author[1]{\textbf{Hunter Heidenreich}}
\author[1]{\textbf{Ratish Dalvi}}
\author[1]{\textbf{Rohith Mukku}}
\author[1]{\textbf{Nikhil Verma}}
\author[1]{\textbf{Neven Pičuljan}}
\affil[1]{Roots Automation, New York, NY}
\affil[ ]{\texttt{ai@rootsautomation.com}}

\begin{document}
\maketitle

\begin{abstract}
    Page Stream Segmentation (PSS) is an essential prerequisite for automated document processing at scale.
    However, research progress has been limited by the absence of realistic public benchmarks.
    This paper works towards addressing this gap by introducing \href{https://huggingface.co/datasets/rootsautomation/TABMEpp}{TABME++}, an enhanced benchmark featuring commercial Optical Character Recognition (OCR) annotations. 
    We evaluate the performance of large language models (LLMs) on PSS, focusing on decoder-based models fine-tuned with parameter-efficient methods. 
    Our results show that decoder-based LLMs outperform smaller multimodal encoders. 
    Through a review of existing PSS research and datasets, we identify key challenges and advancements in the field. 
    Our findings highlight the key importance of robust OCR, providing valuable insights for the development of more effective document processing systems.
\end{abstract}

\section{Introduction}

Digital documentation pervades all aspects of modern business transactions, archival and library sciences, and overall contemporary information management.
The proliferation of digital documents presents novel challenges, such as the labor-intensive process of bulk uploading and ingestion. 
Often, disparate documents are bundled together, without proper delineation, to streamline operations or meet functional needs. 
This lack of clear separation can cause inefficiencies and difficulties in document retrieval and processing.

Automated document processing carries significant risk and financial implications, especially in legal and healthcare domains. Therefore, precise classification and information extraction are paramount. Page stream segmentation (PSS)\footnote{
Historically referred to as document separation, document flow segmentation, document stream segmentation, document bundle separation, and page stream separation.
We use \emph{page stream segmentation} (PSS) for the separation of page sequences into atomic documents. 
We refer to collections of pages as \emph{streams} throughout this work.
} \cite{collins_thompson_clustering_based_2002,gordo_document_2013} plays a critical role in achieving precision by segmenting a sequence of pages into atomic documents, enabling accurate classification and specialized information extraction. 
Low-quality segmentation can render classification and extraction impossible, necessitating meticulous automated solutions.

Despite the importance of segmentation in automated document processing pipelines, prior work on PSS has been severely limited by the lack of public benchmarks that reflect realistic document processing scenarios.
Most studies use private datasets or unrealistic public benchmarks, making comparisons and improvements difficult.

Additionally, large language models (LLMs) exceeding the billion-parameter scale have not been applied to PSS, even though they have been found to perform strongly on document processing benchmarks such as DocVQA \cite{Mathew_2021_WACV,wang_layout_2023,borchmann_notes_2024}.
Instead, most prior work has focused on the application of significantly smaller encoder-based Transformer models \cite{vaswani_attention_2017,devlin_bert_2019} augmented with convolutional neural networks (CNNs).

To address gaps in the PSS literature, we (1) summarize prior research along the dimensions of algorithms, data, and formulations, (2) enhance the TABME benchmark \cite{mungmeeprued_tab_2022,TABME} with commercial-quality OCR, releasing TABME++\footnote{\url{https://huggingface.co/datasets/rootsautomation/TABMEpp}}, (3) evaluate uni- and multimodal encoder models on this benchmark, examining the effects of different modalities, and (4) demonstrate the effective and efficient application of decoder-based LLMs to PSS, showing their superiority over smaller multimodal encoders, thereby effectively solving TABME with quality OCR annotations.

\section{Related Work}

\subsection{Algorithms}

Early PSS systems relied on handcrafted rules, enhanced by machine learning components \cite{collins_thompson_clustering_based_2002, meilender_segmentation_2009, couasnon_document_2013, daher_multipage_2014, karpinski_combination_2016, hamdi_feature_2018}. These methods used region-specific pattern matching on headers and footers, incorporated domain knowledge to identify names, dates, and invoice numbers, and applied rules to partition pages. However, these systems often struggled with heterogeneous documents, lacking generalization capabilities \cite{meilender_segmentation_2009, couasnon_document_2013}.

The first entirely learning-based PSS models used traditional classification or Markov chain models, integrated with pre-deep learning NLP or CV pipelines \cite{Schmidtler2007, gordo_document_2013, rusinol_multimodal_2014, agin_approach_2015}. Notably, \cite{rusinol_multimodal_2014} was the first to explore multimodal PSS, demonstrating effective combinations of textual and visual features. With the advent of word vectors \cite{mikolov_distributed_2013,mikolov_efficient_2013}, there was a paradigm shift toward neural methods in NLP. Researchers achieved success using models like doc2vec \cite{hamdi_machine_2017, neche_use_2020}, CNNs \cite{gallo_deep_2016, wiedemann_multi_modal_2021, BRAZ2021104394, arif_demirtas_semantic_2022, busch_using_2023}, and RNNs \cite{neche_use_2020}.

Despite the recent prominence of Transformer architectures \cite{vaswani_attention_2017} in NLP and beyond, only two studies have applied Transformers to PSS. \cite{guha_multimodal_2022} uses LEGAL-BERT \cite{chalkidis_legal_bert_2020} combined with a CNN. Similarly, \cite{mungmeeprued_tab_2022} combines LayoutLM \cite{xu_layoutlm_2020} with a CNN. This raises questions about Transformer performance on PSS, especially with respect to input modalities. Moreover, the use of decoder-based LLMs has yet to be explored.

\subsection{Datasets}

Research on PSS has been limited by the lack of public datasets due to privacy concerns like PII or trade and financial secrets. This issue extends beyond PSS, affecting many document-based tasks \cite{van_landeghem_beyond_2024}. PSS, being fundamentally multi-page, requires both single and multi-page documents, pre-arranged in streams or suitable for synthetic arrangement, with care to avoid creating unrealistic datasets.

A few public datasets exist in Portuguese \cite{BRAZ2021104394, luz_de_araujo_sequence_aware_2023} and Dutch \cite{van_heusden_wooir_2022}, but only two are available in English:
\begin{itemize}
    \item \textbf{Tobacco800} \cite{CDIP, SIGIR-06, tobacco800}: A subset of the Legacy Tobacco Document Library (LTDL) \cite{LTDL} with fewer than 800 scanned documents made public during legal proceedings against U.S. tobacco companies. Artificial streams are generated for analysis, with prior researchers often approaching it as a page-pair binary classification task, disregarding in situ performance effects on downstream predicted documents.
    \item \textbf{TABME} \cite{mungmeeprued_tab_2022,TABME}: A subset of the Truth Tobacco Industry Documents archive, the modern LTDL. Documents are sampled, placed in train/validation/test splits, and arranged in synthetic streams \textit{with replacement}.
\end{itemize}

\subsection{Problem Formulations}

When designing a PSS system, two components need formulation: the input context and the output target. Most studies use pairs of pages as input (previous and pivot pages), predicting whether there is a break between them, while some classify isolated pages \cite{guha_multimodal_2022} or centered windows of pages \cite{meilender_segmentation_2009,couasnon_document_2013,gordo_document_2013,mungmeeprued_tab_2022}, treating the problem as a binary decision over whether the center page starts a new document. However, \cite{Schmidtler2007} considers a ternary classification problem with beginning, middle, and end classes assigned to each page. Alternatively, \cite{BRAZ2021104394} considers a tertiary classification task with classes such as first of many, not first not last, last page, and single-page document. These class designations can be reduced to binary decisions over pages, reducing to the original formulation without loss of information.

\section{Data}

Before training models to perform PSS, we first analyze public datasets like TABME and Tobacco800, comparing them with private in-house data.
We examine Tobacco800, TABME, TABME++, and an internal dataset of heterogeneous streams collected from the insurance industry, discussing their distributions of streams, documents, pages, and tokens.

\subsection{Extending TABME: TABME++}

Initially, the TABME dataset was used as published. However, the OCR provided by the open-source Tesseract library \cite{TessOverview} introduced significant noise in the detected text. Many pages, especially those with scanning artifacts, were detected as blank, missing valuable features like titles and ID numbers. Given the critical importance of OCR quality \cite{biten_icdar_2019,mafla_multi_modal_2021,biten_ocr_idl_2022}, particularly for unimodal text-based models, we reprocessed every document in TABME using Microsoft OCR\footnote{\url{https://learn.microsoft.com/en-us/azure/ai-services/computer-vision/overview-ocr}}. The improved annotations are released as TABME++.

\begin{figure}[htbp]
\centering
\begin{minipage}[b]{0.4\textwidth}
    \centering
    \includegraphics[width=\linewidth]{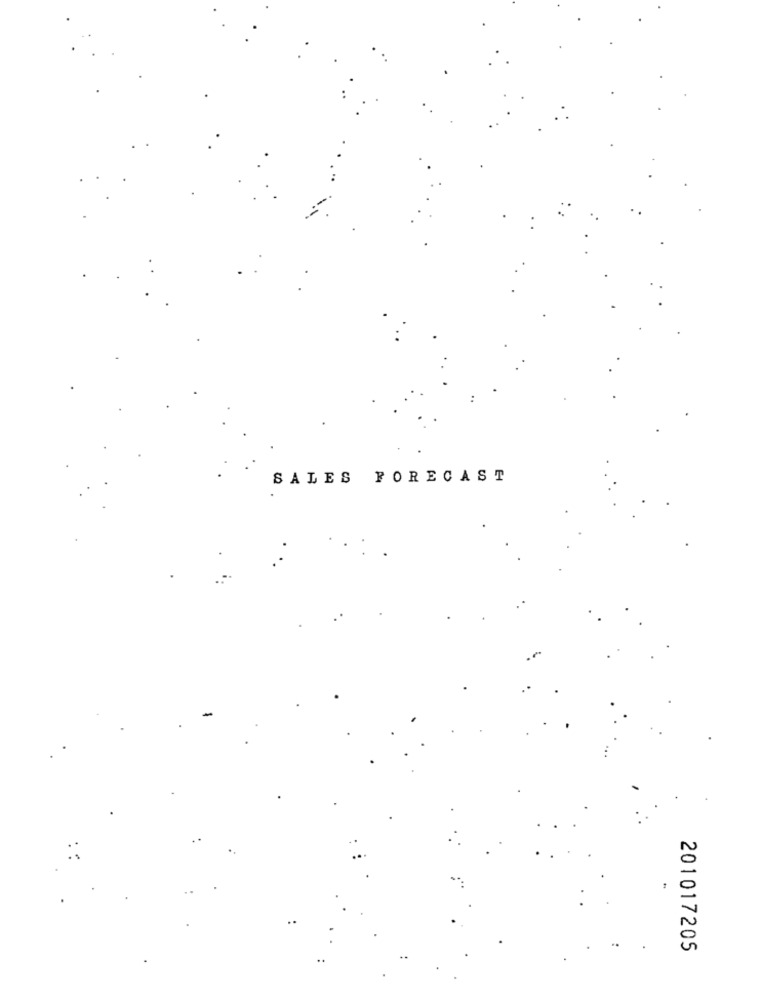}
\end{minipage}%
\hspace{0.1\textwidth}%
\begin{minipage}[b]{0.4\textwidth}
    \centering
    \begin{lstlisting}[style=lineartext, label=lst:texttsv]
02Z10102
    \end{lstlisting}
    \vspace{2.5cm}
    \begin{lstlisting}[style=lineartext, label=lst:text2d]
                SALES FORECAST
                                201017205
    \end{lstlisting}
\end{minipage}
\caption{Example page (document ID: \texttt{ffyw0199}, page ID: 7) illustrating the importance of high-quality OCR. The left side shows the page image, while the right side compares the original OCR (top) with the improved OCR (bottom). Text is projected in 2D to preserve layout, which benefits LLM processing \protect\cite{wang_layout_2023,li_large_2024,bayani_testing_2024}.}
\label{fig:ocr_quality}
\end{figure}

\begin{table}[htbp]
    \centering
    \begin{tabular}{ccc}
    \toprule
         & TABME & TABME++ \\
    \midrule
    Blank Pages & 2,785 (2.27\%) & 466 (0.38\%) \\
    \bottomrule
    \end{tabular}
    \caption{Comparison of the number of blank pages between TABME and TABME++.}
    \label{tab:tabme_blanks}
\end{table}

Figure \ref{fig:ocr_quality} illustrates an improved sample where Tesseract misses a section title and misreads an ID number, while Microsoft OCR accurately detects both. 
Another key improvement is the reduced number of blank pages, as shown in Table \ref{tab:tabme_blanks}.
Previously, about 2\% of the dataset was blank whereas less than half a percent is after reprocessing, indicating that many of these pages had text missed by Tesseract.

\subsection{Cross-Dataset Comparison}

\begin{table}[htbp]
\centering
\begin{tabular}{>{\centering\arraybackslash}m{2cm} r r r r r r r r r}
    \toprule
    \multirow{2}{*}{\textbf{Dataset}} & \multicolumn{2}{c}{\textbf{Streams}} & \multicolumn{3}{c}{\textbf{Documents}} & \multicolumn{2}{c}{\textbf{Pages}} & \multicolumn{2}{c}{\textbf{Tokens}}  \\
    & Total & \% Single & Total & Unique & \% Single & Total & Unique & Total & Unique \\
    \midrule
    Tobacco800 & 1 & 0.00 & 742 & 742 & 66.85 & 1.3K & 1.3K & 935.2K & 934.6K \\
    TABME & 110.0K & 0.02 & 1.2M & 44.8K & 55.35 & 3.3M & 122.5K & 719.1M & 26.6M \\
    TABME++ & 110.0K & 0.02 & 1.2M & 44.8K & 55.35 & 3.3M & 122.5K & 9.5B & 349.7M \\
    Internal & 6.0K & 55.68 & 12.3K & 12.3K & 46.06 & 30.3K & 30.3K & 116.4M & 116.4M \\
    \bottomrule
\end{tabular}
\caption{
    Numerical statistics for datasets, with token counts computed using the GPT-2 tokenizer \protect\cite{radford2019language}. `Unique' refers to counts based on unique documents.
}
\label{tab:dataset_stats}
\end{table}

Table \ref{tab:dataset_stats} provides statistics on streams, documents, pages, and tokens for each dataset. Detailed statistics, including KDE plots, are provided in Appendix \ref{app:kde}.

TABME looks different from our internal dataset, having many streams but few single-document streams. Effective PSS on the internal dataset requires a model to recognize when not to segment a stream, a skill TABME may not develop due to its lack of single-document streams. This discrepancy could complicate transfer learning between the datasets.

At the document level, the datasets are more similar. While TABME has a higher proportion of single-page documents, singletons are common across all datasets. Notably, the internal dataset has the lowest percentage of singletons, indicating that TABME is more representative than Tobacco800, where most documents are singletons.

The unique documents metric highlights the re-sampling in TABME. If re-sampling were uniform (which it was not), each document would be sampled roughly 27 times. This means a single pass through the dataset would expose a model to the same document multiple times.

In terms of tokens, TABME has fewer tokens than the internal dataset despite having four times the number of pages. Post-reprocessing, TABME++ has about three times the number of tokens, aligning its token distribution more closely with the internal dataset, indicating improved OCR quality.

\section{Page Stream Segmentation (PSS)}

\subsection{Task Description}

Given a sequence of $N$ pages, $P = (p_1, p_2, ..., p_N)$, a model is tasked with inferring the document boundaries to generate a new sequence of $M$ documents, $D = (d_1, ..., d_M)$, where $d_k = (p_i, p_{i+1}, ..., p_j) \sqsubseteq P$ is a contiguous sub-sequence of $P$. PSS is treated here as a binary classification problem. For each page, the model classifies whether it starts a new document (positive) or continues the current document (negative). Consequently, for a sequence of $N$ pages, the model outputs a binary $N$-dimensional vector $\hat{\mathbf{y}} \in \mathbb{Z}_2^N$, which dictates the predicted division of documents.

Note that since each page $p_i$ is associated with a binary outcome $y_i$, prediction can be performed by using each page as a center and looking at a window of adjacent pages $(p_{i-l}, ..., p_{i-1}, p_i, p_{i+1}, ..., p_{i+r}) \mapsto y_i$.
Throughout this work, $l=1, r=0$, and so we consider learning classification over page pairs.

Similar to \cite{mungmeeprued_tab_2022}, this formulation assumes all pages in a single document are contiguous and prearranged in sequence. Without these assumptions, we face the broader challenge of clustering page sets into documents without a predefined order, which is atypical for our intended application.

\subsection{On Sampling with Replacement}

Segmentation tasks, including PSS and general text segmentation, are often imbalanced problems. Typically, the number of negative samples (non-breaks) far exceeds the number of positive samples (breaks) \cite{koshorek_text_2018,retkowski_text_2024}.
When operating on page pairs, only documents with exactly two pages evenly contribute to both positive and negative classes when placed in a stream.
Singleton documents bias towards the positive class; documents with more than two pages bias towards the negative.

We can analyze this bias if we make several simplifying assumptions.
We will assume a dataset of $M$ streams and that all streams are independent of each other.
Additionally, we assume that the number of documents in a stream $L_m$ is a random variable following a Poisson distribution with parameter $\lambda$,
that document page length $N_{m,\ell}$ is a random variable following a Poisson distribution with parameter $\nu$, and that these random variables are independent.
Furthermore, we make the naive simplification that documents are fully independent of one another.
Since the mean of a Poisson distribution is equal to its parameter value, in expectation, these assumptions give a dataset of $M\lambda\nu$ pages.

First, consider the scenario where all documents are unique. 
Each document contributes a single positive sample for its start and $\nu - 1$ negative samples for its remaining pages, resulting in a positive class ratio of $\frac{M \lambda}{M \lambda \nu} = \nu^{-1}$.
This aligns with the intuition that the longer documents are, the more negatively imbalanced the dataset will be.

In contrast, for synthetically constructed datasets (like TABME), documents are sampled with replacement from a fixed collection.
For such a dataset with a ratio $p \in (0, 1)$ duplicate documents, one either accepts non-unique data or deduplicates the data to remove repeat samples.
If one accepts the duplicate data, the dataset size remains constant and identical class imbalance exists.

However, if one deduplicates the data, one retains a lower-bound of $(1-p)$\% of the dataset and alters the positive sample ratio to $p + (1-p)\nu^{-1}$ because, for resampled documents, only positive samples are retained.
This has an interesting effect where, in the case of significant resampling, the problem imbalance is shifted entirely towards the positive class.

\begin{table}[htbp]
    \centering
    \begin{tabular}{c c c c c c}
        \toprule
        & \textbf{Resample?} & \textbf{Deduplicate?} &  \textbf{Samples} & \textbf{Unique Ratio} & \textbf{Positive Ratio} \\ 
        \midrule
        Abstract & X & X & $M\lambda\nu$ & 1.00 & $\nu^{-1}$\\
        Abstract & \checkmark & X & $M\lambda\nu$ & $1-p$ & $\nu^{-1}$\\ 
        Abstract & \checkmark & \checkmark & $(1-p)M\lambda\nu$ & 1.00 & $p + (1-p)\nu^{-1}$\\ 
        \midrule
        TABME++ & \checkmark & X & 3,316,922 & 0.36 & 0.3650 \\
        TABME++ & \checkmark & \checkmark & 1,189,590 & 1.00 & 0.9341 \\
        \bottomrule
    \end{tabular}
    \caption{
    Sample counts, unique data ratios, and positive class ratios for datasets under varying assumptions. Here, $M$ denotes the number of streams, $\lambda$ represents the average number of documents per stream, $\nu$ is the average number of pages per document, and $p$ indicates the ratio of duplicated data.
    }
    \label{tab:resample}
\end{table}

We summarize these observations in Table \ref{tab:resample}, and additionally report the empirical values measured from TABME++.
While these assumptions are overly simplistic, they provide insights into the PSS class imbalance problem, illustrating the potential side-effects of synthetic construction.
For our experiments, we use deduplicated data, re-weighting the positive class to help with imbalance in all experiments, except for those with decoder-based LLMs.

\subsection{Evaluation Metrics}

\subsubsection{Page-Level}

For per-page binary classification, we use precision, recall, and F1 score. Although segmentation-specific metrics such as P$_k$ \cite{beeferman_text_1997} and WindowDiff \cite{pevzner_critique_2002} exist for generalized text segmentation, we prioritize document-centric metrics that better align with practical applications of a PSS system.

\subsubsection{Document-Level}

To evaluate performance at the document level, each page is uniquely numbered, and documents are represented as tuples of these sequential numbers (e.g., \(d_k = (p_i, p_{i+1}, \ldots, p_j)\)). For a predicted segmentation \(\mathcal{P} = \{\hat{d}_i\}_{i=1}^{N}\) and a ground truth segmentation \(\mathcal{G} = \{d_j\}_{j=1}^{M}\), we define:

\begin{itemize}
    \item \textbf{True Positives (TP)}: Documents present in both \(\mathcal{P}\) and \(\mathcal{G}\): $|\mathcal{P} \cap \mathcal{G}|$
    \item \textbf{False Positives (FP)}: Documents present in \(\mathcal{P}\) but not in \(\mathcal{G}\): $|\mathcal{P} \setminus \mathcal{G}|$
    \item \textbf{False Negatives (FN)}: Documents present in \(\mathcal{G}\) but not in \(\mathcal{P}\): $|\mathcal{G} \setminus \mathcal{P}|$
\end{itemize}

Document-level precision, recall, and F1 score are then computed as follows:
\begin{align*}
    \text{P} = \frac{|\mathcal{P} \cap \mathcal{G}|}{|\mathcal{P} \cap \mathcal{G}| + |\mathcal{P} \setminus \mathcal{G}|}, \quad 
    \text{R} = \frac{|\mathcal{P} \cap \mathcal{G}|}{|\mathcal{P} \cap \mathcal{G}| + |\mathcal{G} \setminus \mathcal{P}|}, \quad 
    \text{F1} = \frac{2 \cdot \text{P} \cdot \text{R}}{\text{P} + \text{R}}
\end{align*}

These metrics provide a comprehensive measure of prediction quality at the document level, focusing on the correctness and completeness of the predicted document boundaries.

\subsubsection{Stream-Level}

At the stream level, we adopt the Minimum Number of Drag-and-Drops (MNDD) metric from \cite{mungmeeprued_tab_2022}, representing the number of pages a human would need to drag and drop to correct the prediction, allowing for document-ordering to be shuffled. We also employ the straight-through processing (STP) metric, the ratio of streams the model segments perfectly, requiring no human intervention.

\subsection{Methods}

\subsubsection{Bidirectional Encoders}

Prior work using Transformer-based models for PSS has focused on bidirectional encoder models, often augmented with vision-based CNNs \cite{guha_multimodal_2022,mungmeeprued_tab_2022}. Previous studies have considered both single-page and multi-page inputs. We standardize encoder inputs by presenting each model with two pages and tasking it with predicting continuity or breakage.

We aim to understand how modality influences the quality of an encoder-based PSS model. We consider a cross-sampling of encoders, each with different input modalities: RoBERTa (text-only) \cite{devlin_bert_2019,liu_roberta_2019}, DiT (vision-only) \cite{li_dit_2022}, LiLT (text and layout) \cite{wang_lilt_2022}, and LayoutLMv3 (text, vision, and layout) \cite{huang_layoutlmv3_2022}.

Encoder models have shorter context lengths $L$ than their decoder-based counterparts ($L=512$ tokens for all models considered here). To address this limitation, models with text-based inputs encode both the first $L$ and last $L$ tokens as separate vectors, concatenating them as a single representation before classifying the page pair as a break or continuity:

\begin{align}
    \mathbf{\hat{y}_i} = W \left(E(p_{i-1}) \oplus E(p_i) \right) + \mathbf{b}
\end{align}

\begin{equation}
    E(p_j) =
    \begin{cases}
      E(p_{j, :L}) \oplus E(p_{j, -L:}), & \text{if text-based} \\
      E(p_j), & \text{otherwise}
    \end{cases}
\end{equation}

where $E$ represents the encoder and $E(\cdot)$ the encoding process of a page, $\oplus$ represents the concatenation of the vector embeddings of the two pages, and $W \in \mathbb{R}^{2 \times 2d}, \mathbf{b} \in \mathbb{R}^2$ are learned parameters for a 2D classification output layer that predicts based on $d$-dimensional encoder outputs. 
While prior works, such as \cite{mungmeeprued_tab_2022}, have opted for more complex prediction layers such as 1D convolution on the sequence of page representations to produce a sequence of predictions, we opt for a simpler prediction layer to focus on the core of the page-pair binary classification problem.
Training and hyperparameter details can be found in Appendix \ref{app:hyper}.

\subsubsection{Autoregressive Decoders}

To explore the application of decoder-only LLMs for PSS, we consider two open-weight models: Mistral-7B \cite{jiang_mistral_2023} and Phi-3-mini \cite{abdin_phi_3_2024}, with model sizes of 7 billion and 4 billion parameters, respectively. For parameter-efficient fine-tuning (PEFT), we employ Low-Rank Adaptation (LoRA) \cite{hu_lora_2021}. Training and hyperparameter details for each experiment are in Appendix \ref{app:hyper}.
For all decoder models, we use a standardized task description prompt, structuring it in the model's preferred format using its special tokens. This template is shown in Listing \ref{lst:prompt} in Appendix \ref{app:hyper}.

Models output a simple JSON object with a single key-value pair containing their prediction. If the JSON is malformed or lacks the proper key, the model defaults to predicting $0$ (current page continues the previous document). We also report zero-shot performance for Mistral-7B, Phi-3, and GPT-4o\footnote{Model version \texttt{2024-05-13}} using the same prompt to establish performance prior to fine-tuning.
For GPT-4o, we test on a random 10\% subset of test set streams to keep costs of evaluation reasonable.

\subsubsection{Baseline}

As a traditional ML baseline, we use an XGBoost model \cite{chen_xgboost_2016} trained on the concatenation of count- and TF-IDF-based representations. This bag-of-words model lacks the ability to contextualize the statistics it ingests but benefits from a global, page-level view without context length limitations.

\section{Results}
\label{sec:results}

\subsection{TABME++}

\begin{table}[htbp]
\centering
\begin{tabular}{>{\centering\arraybackslash}m{2.5cm} r r r r r r r r r r}
    \toprule
    \multirow{2}{*}{\textbf{Model}} & \multicolumn{3}{c}{\textbf{Page}} & \multicolumn{3}{c}{\textbf{Document}} & \multirow{2}{*}{\textbf{MNDD} $\downarrow$} & \multirow{2}{*}{\textbf{STP} $\uparrow$} \\
    \cmidrule(lr){2-4} \cmidrule(lr){5-7}
    & \textbf{Precision} & \textbf{Recall} & \textbf{F1 Score} & \textbf{Precision} & \textbf{Recall} & \textbf{F1 Score} & & \\
    \midrule
    \textbf{Traditional} \\
    XGBoost & 0.806 & 0.859 & 0.831 & 0.616 & 0.670 & 0.642 & 10.852 & 0.074 \\
    \midrule
    \textbf{Encoders} \\
    RoBERTa & 0.666 & 0.932 & 0.777 & 0.479  & 0.671 & 0.559 & 12.169 & 0.042 \\
    DiT & 0.794 & 0.871 & 0.831 & 0.596 & 0.654 & 0.623 & 10.479 & 0.066 \\
    LiLT & 0.648 & 0.927 & 0.763 & 0.457 & 0.657 & 0.539 & 12.863 & 0.037 \\
    LayoutLMv3 & 0.672 & 0.942 & 0.784 & 0.490 & 0.687 & 0.572 & 11.329 & 0.048 \\
    \midrule
    \textbf{Decoders} \\
    Phi-3-mini-ZS  & 0.370 & 0.147 & 0.211 & 0.092 & 0.044 & 0.060 & 14.785 & 0.000 \\
    Mistral-7B-ZS  & 0.426 & 0.884 & 0.575 & 0.270 & 0.578 & 0.368 & 21.146 & 0.014 \\
    Phi-3-mini-FT & 0.955 & 0.991 & 0.973 & 0.916 & 0.955 & 0.933 & 1.759 & 0.637 \\
    Mistral-7B-FT  & \textbf{0.975} & \textbf{0.999} & \textbf{0.987} & \textbf{0.955} & \textbf{0.979} & \textbf{0.967} & \textbf{0.811} & \textbf{0.800} \\
    \cmidrule(lr){2-9}
    GPT-4o-ZS$^\ast$ & 0.763 & 0.982 & 0.859 & 0.624 & 0.805 & 0.703 & 7.794 & 0.094 \\
    \bottomrule
\end{tabular}
\caption{Model performance comparison on TABME++. Models are categorized by type, with ZS representing zero-shot performance and FT indicating fine-tuned models. Metrics are evaluated on the test set. GPT-4o's performance is based on 10\% of the test set, providing an approximate measure of its true capability. The best value in each column is highlighted in bold.}
\label{tab:model_comparison_tabme++}
\end{table}

We train models on TABME++ for a maximum of three epochs. 
To monitor training performance, we down-sample the validation split to 10\% of the streams and use the page-level validation F1 to select the best checkpoint. 
After training and model selection, all models are evaluated on the full test split, with metrics displayed in Table \ref{tab:model_comparison_tabme++}.

Overall, we find that fine-tuned decoder-based LLMs can effectively learn PSS on TABME++, achieving an F1 score greater than 0.9 at the document level.
The best model, Mistral, achieves an STP of 0.8, meaning it perfectly segments 80\% of streams in the test set. 
This contrasts sharply with their zero-shot counterparts which perform poorly on PSS, including higher-end, closed-source models like GPT-4o.
Nonetheless, GPT-4o outperforms all encoder and traditional models, which is impressive given its lack of tuning in this domain.

Encoder models fail to outperform an XGBoost baseline, even with extended fine-tuning. 
This performance gap might be mitigated by further hyperparameter tuning. 
For example, \cite{mungmeeprued_tab_2022} fine-tune for 10 epochs. 
However, performing such a parameter search would be computationally expensive and hard to justify given the success decoder-only models achieve without such sensitivity and while observing only a fraction of the training samples.

To highlight the sample efficiency of the decoder models, we present the training losses of Mistral-7B and Phi-3-mini in Figure \ref{fig:sample_eff}.
As reported in Appendix \ref{app:hyper}, we continue training both decoders for 20,000 updates after which we find validation metrics begin to decrease.

\begin{figure}[h]
    \centering
    \begin{minipage}[t]{0.4\textwidth}
        \centering
        \includegraphics[width=\linewidth]{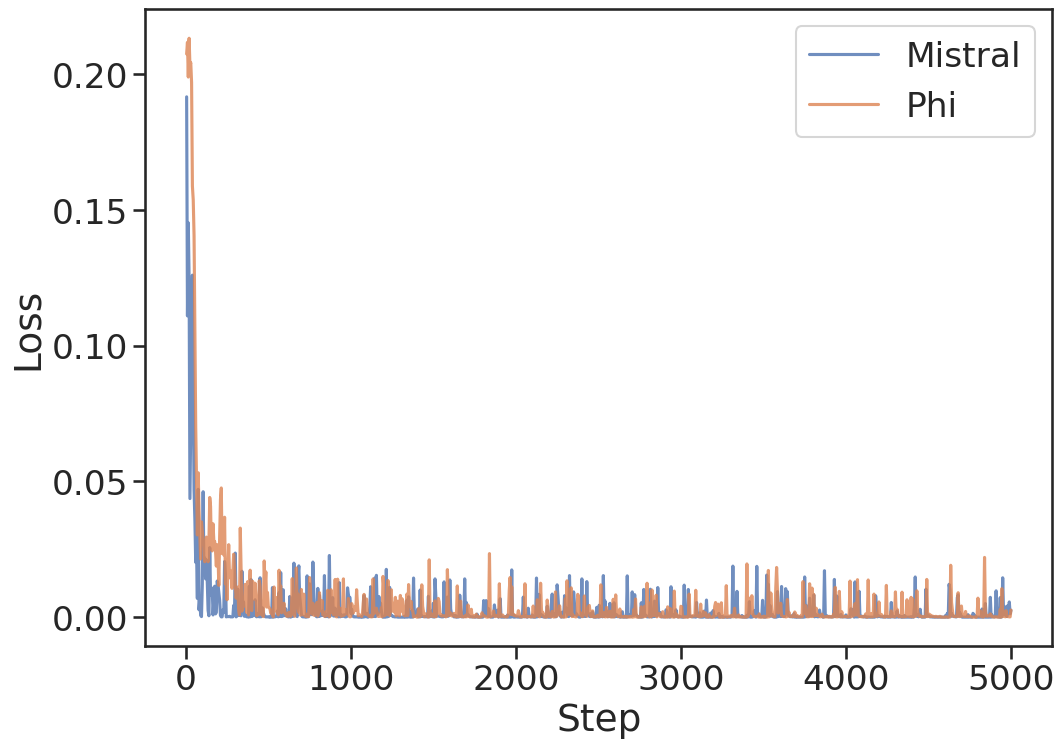}
    \end{minipage}%
    \hfill
    \begin{minipage}[t]{0.6\textwidth}
        \centering
        \vspace{-100pt} % Aligns the top of the table with the image
        \begin{tabular}{>{\centering\arraybackslash}m{2.0cm} r r r r r r r r}
            \toprule
            \multirow{2}{*}{\textbf{Model}} & \multicolumn{3}{c}{\textbf{Page}} & \multicolumn{3}{c}{\textbf{Document}} \\
            \cmidrule(lr){2-4} \cmidrule(lr){5-7}
            & \textbf{P} & \textbf{R} & \textbf{F1} & \textbf{P} & \textbf{R} & \textbf{F1} \\
            \midrule
            Phi-3-mini & \textbf{0.956} & 0.999 & \textbf{0.977} & \textbf{0.929} & \textbf{0.971} & \textbf{0.950} \\
            Mistral-7B & 0.950 & \textbf{1.000} & 0.974 & 0.909 & 0.957 & 0.932 \\
            \bottomrule
        \end{tabular}
    \end{minipage}
    \caption{Sample efficiency of decoder-based LLMs, demonstrating rapid convergence within the first 1,000 updates. Validation metrics for weights after 5,000 updates are presented, indicating robust performance early in training. The best in each column is highlighted in bold.}
    \label{fig:sample_eff}
\end{figure}

To further understand the performance differences among encoder models, we examine the impact of various input modalities, including text-only, vision-only, and multimodal approaches, on their effectiveness in PSS tasks.
DiT, which uses vision-only input, is the strongest encoder model. 
Surprisingly, LayoutLMv3, which integrates vision, text, and layout, does not surpass DiT's performance. 
Additionally, the LiLT model performs slightly worse than the unimodal RoBERTa model, despite starting from the same base weights and incorporating layout information. 
This suggests that the inclusion of additional modalities does not necessarily improve performance on this dataset and that further investigation is needed to understand the interplay between these modalities and model architecture.

\subsubsection{TABME}

To better understand the effects of augmenting TABME with commercial quality OCR annotations, we also train a subset of unimodal models on the original TABME annotations.
We show the results of these experiments alongside previously reported performance in Table \ref{tab:model_comparison_tabme}.

\begin{table}[htbp]
\centering
\begin{tabular}{>{\centering\arraybackslash}m{4.5cm} r r r r r r}
    \toprule
    \multirow{2}{*}{\textbf{Model}} & \multicolumn{3}{c}{\textbf{Page-Level}} & \multirow{2}{*}{\textbf{MNDD} $\downarrow$} \\
    \cmidrule(lr){2-4}
    & \textbf{Precision} & \textbf{Recall} & \textbf{F1 Score} & \\
    \midrule
    \cite{mungmeeprued_tab_2022} & 0.947 & 0.964 & 0.953 & 3.440 \\
    \cite{mungmeeprued_tab_2022}, ResNet-only & 0.944 & 0.948 & 0.942 & 3.910 \\
    \cite{mungmeeprued_tab_2022}, LayoutLM-only & 0.940 & 0.421 & 0.559 & 15.928 \\
    \midrule
    XGBoost & 0.762 & 0.784 & 0.773 & 12.595 \\
    RoBERTa & 0.766 & 0.653 & 0.705 & 13.409 \\
    Mistral-7B-ZS & 0.388 & 0.952 & 0.552 & 19.443 \\
    Mistral-7B-FT & 0.920 & 0.940 & 0.930 & 4.369 \\
    \bottomrule
\end{tabular}
\caption{Performance comparison of text-based models on TABME.}
\label{tab:model_comparison_tabme}
\end{table}

When comparing the performance across datasets, XGBoost, RoBERTa, and Mistral all experience a drop of around 0.05-0.07 in page-level F1 compared to their performance on TABME++.
RoBERTa achieves much higher recall on TABME++ than on the prior iteration, TABME.
Fine-tuned Mistral still offers extremely strong performance, outperforming all other text-based models on TABME, and approaching the performance of the ResNet-only and multimodal model state-of-the-art \cite{mungmeeprued_tab_2022}.

Additionally, it's worth emphasizing that models in \cite{mungmeeprued_tab_2022} make predictions about all pages in a stream jointly with a convolutional prediction head whereas all models we consider only observe page pairs.
It may be that given more pages in sequence, Mistral can achieve an even higher performance, but we leave this as an open question.

\section{Conclusion}

In this work, we highlight key challenges in Page Stream Segmentation (PSS) and introduce TABME++, an enhanced benchmark with high-quality OCR annotations. Our evaluations show that fine-tuned decoder-based large language models (LLMs) outperform smaller multimodal encoders and traditional baselines, demonstrating their potential for effective document segmentation.

Our findings highlight the importance of robust OCR in improving PSS accuracy. The enhancements in TABME with commercial-grade OCR led to significant performance gains, underscoring the criticality of accurate text extraction. The analysis of different model architectures provides valuable insights into the strengths and limitations of text-only, vision-only, and multimodal approaches.

In future work, we will explore the generalizability of these results to more realistic datasets. Furthermore, we hope to better understand how to intelligently and efficiently leverage multimodality, especially in a way that is compatible with our decoder-only approach.

%Bibliography
\printbibliography

\newpage

\begin{appendices}

\section{KDE Plots}
\label{app:kde}

\begin{figure}[htbp]
    \centering
    \includegraphics[width=0.6\linewidth]{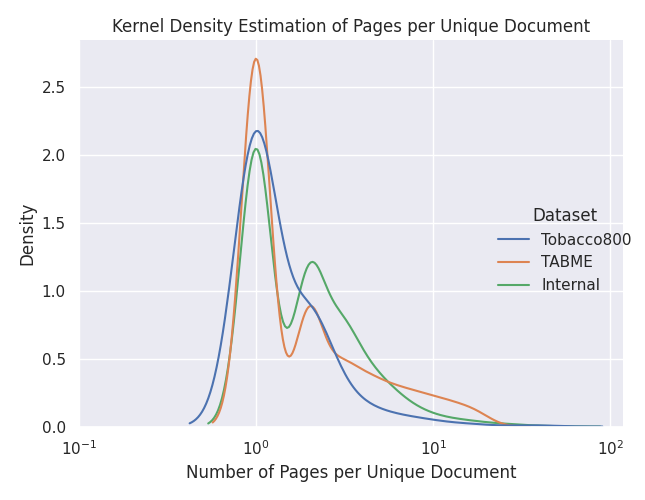}
    \caption{Kernel density estimate (KDE) of the number of pages per document across datasets. The shift from Tobacco800 to TABME highlights a more pronounced bimodal distribution, aligning with our internal dataset but with generally shorter documents compared to TABME, as reflected in the distribution's right-hand tail.}
    \label{fig:kde_pg_doc}
\end{figure}

\begin{figure}[htbp]
    \centering
    \includegraphics[width=0.6\linewidth]{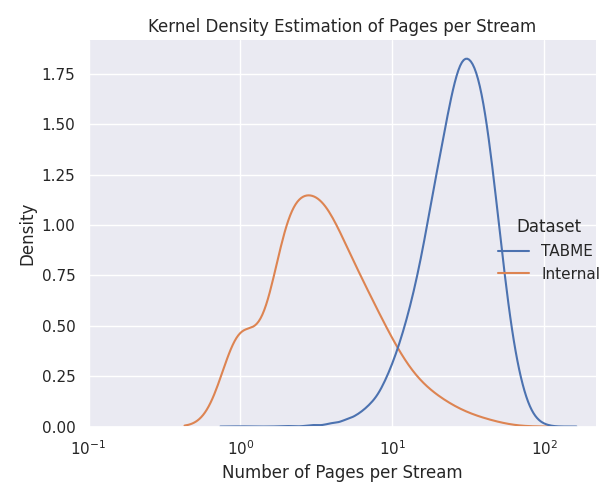}
    \caption{Kernel density estimate (KDE) of the number of pages per stream, showing that TABME features longer streams than those found in our internal dataset.}
    \label{fig:kde_pg_stream}
\end{figure}

\begin{figure}[htbp]
    \centering
    \includegraphics[width=0.6\linewidth]{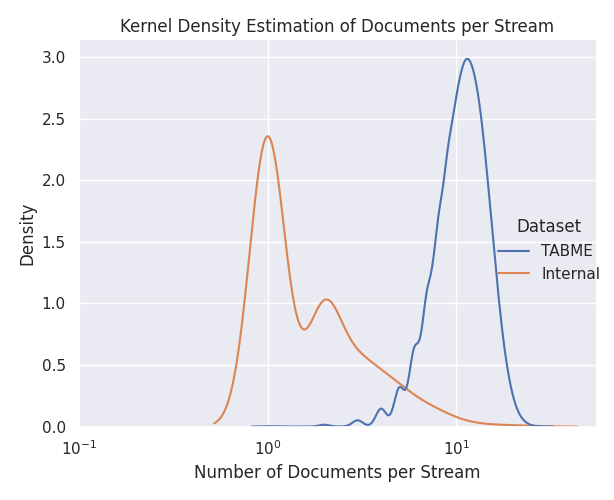}
    \caption{Kernel density estimate (KDE) of the number of documents per stream. TABME tends to have more documents per stream compared to our internal dataset. This suggests a potential need for greater variability in document length when synthetically generating streams or adjusting the Poisson distribution parameters.
    }
    \label{fig:kde_doc_stream}
\end{figure}

\begin{figure}[htbp]
    \centering
    \includegraphics[width=0.6\linewidth]{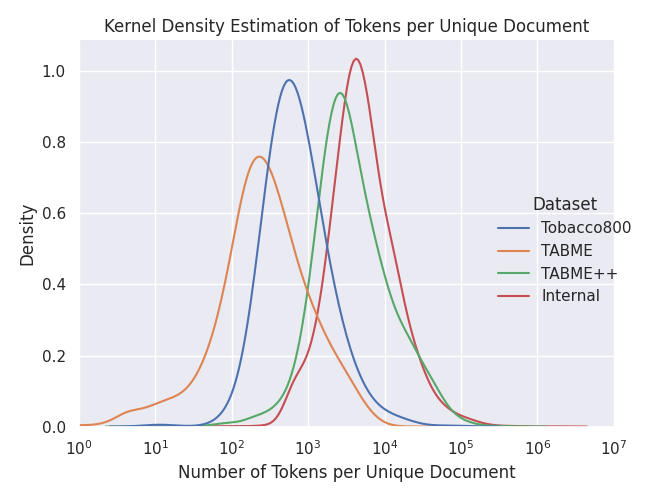}
    \caption{
    Kernel density estimate (KDE) of the number of tokens per document. Enhancing TABME to TABME++ aligns its token distribution more closely with our internal dataset. Prior to this augmentation, TABME was less text-dense than Tobacco800, highlighting issues with the original OCR annotations. Post-augmentation, TABME++ remains slightly less text-dense than our heavily text-laden internal dataset.
    }
    \label{fig:kde_token_docs}
\end{figure}

\begin{figure}[htbp]
    \centering
    \includegraphics[width=0.6\linewidth]{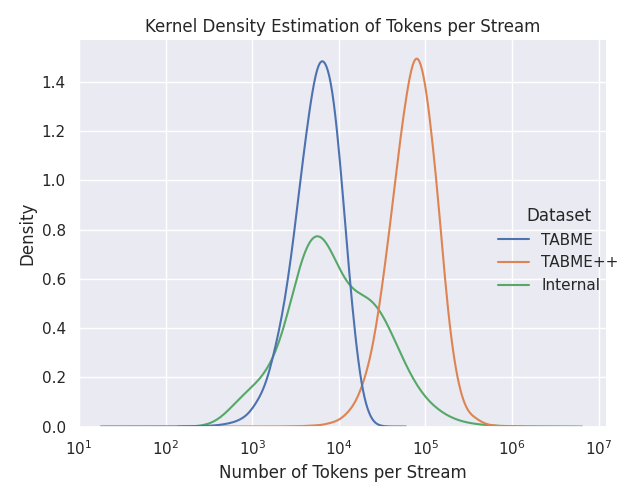}
    \caption{Kernel density estimate (KDE) of the number of tokens per stream. The internal dataset differs significantly in distribution shape, showing a broader variance compared to the concentrated mean of TABME(++) when plotted on a logarithmic scale.
    }
    \label{fig:kde_token_streams}
\end{figure}

\section{Training Details}
\label{app:hyper}

\subsection{Traditional Models}

We employed count-based and TF-IDF-based vector representations using scikit-learn \cite{scikit_learn}. Individual pages were treated as ``documents'' for fitting the vector representations. The text was lowercased, and the default word-based tokenization strategy was applied to produce unigrams. Additionally, TF-IDF vectors were L2-normalized as per the scikit-learn default.

An XGBoost \cite{chen_xgboost_2016} classifier was trained on adjacent page pairs to predict true document breaks. Each page was independently vectorized, and the vectors of the preceding and current pages were concatenated as the input to the XGBoost model. We adjusted for class imbalance by scaling with the positive class ratio and used 100 estimators.

\subsection{Encoders}

Encoder models are fine-tuned using Hugging Face Transformers \cite{wolf2019huggingface}.
We allow backpropagation through all model weights.
Hyperparameters were chosen based on default recommendations from their respective studies, and Hugging Face's training utilities.
We summarize the hyperparameter choices for encoders in Table \ref{tab:encoder_params}.

\begin{table}[htbp]
    \centering
    \begin{tabular}{c|c c c c}
        \toprule
        \textbf{Params.} & RoBERTa & DiT & LiLT & LayoutLMv3 \\
        \midrule
        Peak LR & $10^{-5}$ & $10^{-5}$ & $10^{-5}$ & $10^{-5}$ \\
        Batch size & 32 & 64 & 32 & 32 \\
        LR Warm-up steps & 500 & 500 & 500 & \\
        Weight decay & 0.01 & 0.01 & 0.01 & 0.01 \\ 
        Optimizer & AdamW & AdamW & AdamW & AdamW \\
        Max train epochs & 2 & 2 & 2 & 2 \\ 
        Sequence length & 512x4 & - & 512x4 & 512x4 \\
        Image resolution & - & 224$^2$ & - & 224$^2$ \\ 
        Weighted loss & Yes & Yes & Yes & Yes \\
        \bottomrule
    \end{tabular}
    \caption{Hyperparameters used for fine-tuning encoder models.}
    \label{tab:encoder_params}
\end{table}

\subsection{Decoders}

We primarily rely on Unsloth \cite{unsloth} for performance efficient fine-tuning of LLMs. 
For Mistral-7B, we use the 4-bit quantized version of the instruct v0.2 model\footnote{\url{https://huggingface.co/unsloth/mistral-7b-instruct-v0.2-bnb-4bit}}.
For Phi-3-mini, we use the 4k context length instruct model\footnote{\url{https://huggingface.co/unsloth/Phi-3-mini-4k-instruct}}.
Models are trained using Hugging Face's TRL library \cite{von_Werra_TRL_Transformer_Reinforcement} on completion tokens only, ignoring the instructions when backpropagating.
The prompt template is shown in Listing \ref{lst:prompt}, where \texttt{pg} and \texttt{pg\_prev} attempt to preserve layout structure in 2D \cite{wang_layout_2023}.
All other hyperparameters are summarized in Table \ref{tab:decoder_params}.
We perform all decoder fine-tuning on a single NVIDIA H100 GPU, with LoRA weights in BF16 format.

\begin{lstlisting}[style=prompt, caption={Instruction prompt used for page stream segmentation.}, label=lst:prompt]
You are a skilled document reviewer. Given extracted text from pages of documents, your task is to determine if a page starts a new document or continues from the previous one. You will be presented with the text of the current page and the text of the preceding page.

Example:

Prior text:
###
This is the text on the page before the page you are evaluating.
###
Page text:
###
This is the text on the page you are evaluating.
###

Carefully review the text to decide if the current page starts a new document or continues from the previous one.

Here is the input:

Prior text:
###
{pg_prev}
###
Page text:
###
{pg}
###

Output your prediction as a JSON object. When the page is the start of a new document, your output should be {"label": 1}. If the page continues the document from the previous page, your output should be {"label": 0}. Do not provide any explanation, additional information, or punctuation. Simply provide the JSON object.

Does the page start a new document?
\end{lstlisting}

\begin{table}[htbp]
    \centering
    \begin{tabular}{c|c c}
        \toprule
        \textbf{Params.} & Mistral-7B & Phi-3-mini \\
        \midrule
        Peak LR & $3 \times 10^{-5}$ & $3 \times 10^{-5}$ \\
        Batch size & 16 & 16 \\
        Weight decay & 0.01 & 0.01 \\ 
        Optimizer & \texttt{paged\_adamw\_8bit} & \texttt{paged\_adamw\_8bit} \\ 
        Max train epochs & 3 & 3 \\ 
        Epochs observed & 0.297 (20k updates) & 0.297 (20k updates) \\
        LR warm-up steps & 200 & 200 \\ 
        LoRA $r$ & 16 & 16 \\ 
        LoRA $\alpha$ & 16 & 16 \\ 
        Sequence length & 8192 & 4096 \\ 
        Weighted loss & No & No \\
        \bottomrule
    \end{tabular}
    \caption{Hyperparameters used for PEFT of decoder-only LLMs.}
    \label{tab:decoder_params}
\end{table}

\end{appendices}

\end{document}